%% file: acl_latex.tex
\pdfoutput=1

\documentclass[11pt]{article}

\usepackage[]{acl}

\usepackage{times}
\usepackage{latexsym}
\usepackage{booktabs}
\usepackage{multirow}
\usepackage{algorithmic}%
\usepackage{tcolorbox}
\tcbuselibrary{skins}
\usepackage{xspace}
\usepackage{enumitem}
\usepackage{textcomp}
\usepackage{adjustbox}
\usepackage{wrapfig}
\usepackage{multirow}
\usepackage{amsthm}
\usepackage{subcaption}
\usepackage{comment}
\usepackage[normalem]{ulem}
\usepackage{fancyvrb}
\usepackage{hyperref} 
\usepackage{float}
\usepackage{tabularx} 
\usepackage{caption}
\usepackage{placeins}

\usepackage[T1]{fontenc}

\usepackage[utf8]{inputenc}

\usepackage{microtype}
\usepackage{tabularx}

\title{Evaluating Gender Bias in the Translation \\ of Gender-Neutral Languages into English}

\author{Spencer Rarrick \And Ranjita Naik\thanks{All authors are affiliated with Microsoft.}
\thanks{Contact author at \texttt{\scriptsize ranjitan@microsoft.com}.} \And Sundar Poudel  \And     Vishal Chowdhary
}

\begin{document}
\maketitle
\begin{abstract}

Machine Translation (MT) continues to improve in quality and adoption, yet the inadvertent perpetuation of gender bias remains a significant concern. Despite numerous studies into gender bias in translations from gender-neutral languages such as Turkish into more strongly gendered languages like English, there are no benchmarks for evaluating this phenomenon or for assessing mitigation strategies. To address this gap, we introduce GATE X-E, an extension to the GATE \citep{rarrick2023gate} corpus, that consists of human translations from Turkish, Hungarian, Finnish, and Persian into English. Each translation is accompanied by feminine, masculine, and neutral variants for each possible gender interpretation. The dataset, which contains between 1250 and 1850 instances for each of the four language pairs, features natural sentences with a wide range of sentence lengths and domains, challenging translation rewriters on various linguistic phenomena. Additionally, we present an English gender rewriting solution built on GPT-3.5 Turbo and use GATE X-E to evaluate it. We open source our contributions to encourage further research on gender debiasing.
\end{abstract}

\input{latex/content/1_introduction}
\input{latex/content/2_data_collection}
\input{latex/content/2_5_rewriting_strategy}
\input{latex/content/3_experiments}

\input{latex/content/4_results}

\input{latex/content/5_related_work}
\input{latex/content/6_conclusion}

\bibliography{anthology,custom}
\input{latex/content/7_appendix}
\end{document}

%% file: latex/content/1_introduction.tex
\section{Introduction}

Despite dramatic improvement in general MT quality and breadth of supported languages over recent years \citep{nllbteam2022language}, gender bias in MT output remains a significant problem \citep{piazzolla2023good}. One such type of gender bias is spurious gender-markings in MT output when none were present in the source. This occurs most frequently when translating from a weakly-gendered language into a more strongly gendered one. We explore this phenomenon in translations from Turkish, Persian, Finnish, and Hungarian into English. 

Although English is not a particularly strongly gendered language, it does mark gender through gendered pronouns (\emph{he}, \emph{she}, etc.) and possessive determiners (\emph{his}, \emph{her}), or through a limited number of intrinsically gendered nouns (\emph{mother}, \emph{uncle}, \emph{widow}, etc).

None of our selected source languages use gendered pronouns, but all do use some intrinsically gendered noun words. The exact set of concepts covered by intrinsically gendered words varies per language. For example, Turkish differentiates \emph{mother} (anne) from \emph{father} (baba), but does not differentiate \emph{nephew} from \emph{niece} (both are \emph{yeğen}, though explicit gender-marking modifiers are possible if needed for emphasis). All pronouns in these languages are gender neutral (e.g. Turkish \emph{O} meaning \emph{he/she/singular they}). Additionally in Turkish, subject pronouns and possessive modifiers may be omitted  or appear only in verb and noun morphology. In these cases as well, gender is not marked.

\begin{figure}[t]
\centering \includegraphics[width=0.9\linewidth]{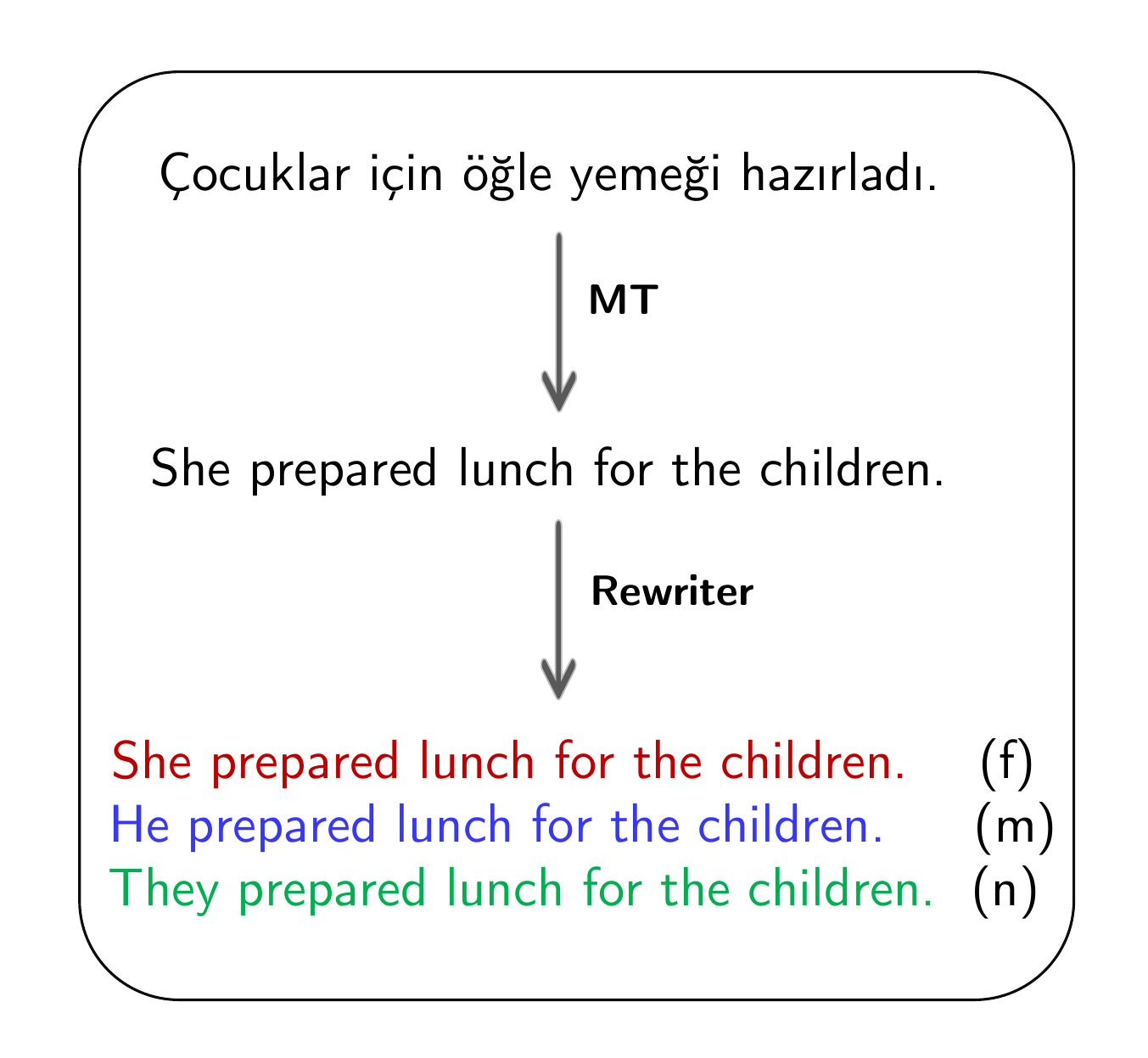}
\caption{ {\bf Gender Bias in Turkish-English Translation.} When translating from Turkish to English, the model tends to use the female pronoun \emph{she} for gender-unspecified individuals, likely due to a perceived link between women and child care. This bias can be mitigated by providing  feminine, masculine, and neutral rewrites.}
\label{fig:mt_bias_mitigation}
\end{figure}

This difference in gender on third-person singular pronouns leads to translation scenarios such as the one seen in Figure \ref{fig:mt_bias_mitigation}, where someone with no specified gender in the source is marked as female in the translation through the pronoun \emph{she}. Machine translation models often make gender assignments according to stereotypes \citep{stanovsky-etal-2019-evaluating} -  in this case a model appears to associate child care with women. One remedy for this category of problems is to supplement the default female translation with masculine and gender-neutral alternatives, covering the range of possible gender interpretations for the source sentence. This could be accomplished by applying a gender rewriter to the original MT output, as shown in the bottom portion of Figure \ref{fig:mt_bias_mitigation}.

GATE (Gender-Ambiguous Translation Examples, \citealt{rarrick2023gate}) presents an evaluation benchmark for gender rewrites for translations from English into French, Spanish, and Italian. In this work, we introduce GATE X-E\footnote{X-E indicates translation from `X' language into English}, an extension to GATE that focuses on translations \emph{into} English from several gender-neutral languages, including Turkish, Persian, Finnish, and Hungarian. It consists of natural sentences with strong diversity of sentence lengths and domains. It challenges translation rewriters on a wide range of linguistic phenomena. GATE X-E contains between 1250 and 1850 instances for each of our language pairs.

We also present a monolingual rewriting algorithm that utilizes GPT-3.5 Turbo \citep{chatgpt} to provide gendered and gender-neutral alternatives for a subset of translation rewrite problems, and achieves high accuracy on the corresponding subset of GATE X-E. We also discuss evaluation and compare to prior work in English gender rewriting. Finally, we perform a detailed error analysis of the results.

The remainder of this paper is organized as follows -- In section \ref{sec:data_collection}, we discuss the corpus creation process for GATE X-E. In section \ref{sec:rewriting_strat}, we discuss strategies for creating gendered alternatives in translations into English. In section \ref{sec:Experiments} we discuss evaluation of our GPT-based rewriting algorithm and those used in prior work on GATE X-E. In section \ref{sec:Results} we discuss results of our experiments and perform detailed error analysis. Finally, In section \ref{sec:Related_Work} we cover related work.

%% file: latex/content/2_data_collection.tex
\section{GATE X-E Dataset}
\label{sec:data_collection}

\subsection{Abitrarily Gender-Marked Entities}

Following \citet{rarrick2023gate}, we use Arbitrarily Gender-Marked Entity (AGME) to refer to individuals whose gender is not marked in the source, but is in the target, either through a gendered pronoun, or an intrinsically gendered noun. Presence of an AGME in a translation indicates that alternate gender translations are possible.

The subject pronoun from the example translation shown in Figure \ref{fig:mt_bias_mitigation} is an example of an AGME. Because there is no gender marking in the source sentence, it is valid to translate the subject as \emph{she}, \emph{he}, or \emph{they}.

\subsection{Dataset Creation and Annotation }

All instances in GATE X-E consist of a single source sentence with one or more translations covering possible gender interpretations. We pulled sentence pairs for each of our language pairs from several corpora found on OPUS\footnote{https://opus.nlpl.eu/}: Europarl, TED talks, tatoeba, wikimatrix, OpenSubtitles, QED and CCAligned. We then apply the following filters:
\begin{itemize}
\item The source sentence scores at least 0.7 match for the intended language when using the python langdetect\footnote{https://pypi.org/project/langdetect/} package. 
\item The English translation contains at least one word on a word list consisting of 79 gendered English nouns (e.g. \emph{mother, uncle, actress, duke}) and pronouns (\emph{he, she, him, her, his, hers, himself, and herself}.)
\end{itemize}

We then sampled sentences from the filtered set and provide them to annotators. From this data, the annotators select appropriate sentences and annotate them for entity types, number of AGMEs, and gendered-alternative translations if AGMEs are present. 

If there are any AGMEs, they will provide a translation as if all AGMEs are female and another as if they are all male. An alternative translation where all AGMEs use gender-neutral language is also provided where possible, so long as it remains an accurate translation of the source sentence: singular \emph{they/them/their/themselves} are used for pronouns, and gender-neutral noun forms are used where they seem reasonably natural at the annotator's judgment. For example, \emph{child} may be used as a gender-neutral variant of \emph{son}, but \emph{nibling} may be deemed too obscure to be used as a variant of \emph{niece}. If exactly two AGMEs are present, we provide a set of variant translations that contain a mixture of female, male, and neutral gender assignments.

Some sentences may contain a mixture of references to AGMEs as well as to humans who are gender-marked in the source. In these cases, gender indicated in the source will be preserved in all translations, as in \emph{father} and \emph{his will} in the example shown in Figure \ref{mix_agmes}. In this example, \emph{Babası} explicitly indicates \emph{father} in the source.

\begin{table}[!h]
\tabcolsep 3pt
\centering
\begin{tabular}{lp{0.4\textwidth}}
\toprule
{\fontsize{10}{12}\bf Src}& {\fontsize{10}{12}Babası vasiyetinde evi ona bıraktı.} \\ 
{\fontsize{10}{12}\bf Fem}&{\fontsize{10}{12}Her father left her the house in his will.} \\  
{\fontsize{10}{12}\bf Masc}&{\fontsize{10}{12}His father left him the house in his will.} \\  
{\fontsize{10}{12}\bf Neut}&{\fontsize{10}{12}Their father left them the house in his will.}  \\ 
{\fontsize{10}{12}\bf Lbls}&{\fontsize{10}{12}target\_only\_gendered\_pronoun, source+target\_gendered\_noun+pronoun,
1-AGME, mixed}  \\ 
\bottomrule
\end{tabular}
\captionof{figure}{{\bf GATE X-E Example Instance.} This includes Turkish source; feminine, masculine and gender-neutral English translations; and labels.}
\label{mix_agmes}
\end{table}

\subsection{Labels}
All Instances in GATE X-E refer to at least one person who is marked for gender in the English target. We include both negative examples, for which the gender marking could be determined from the source, and positive examples, where at least one of those individuals was not gender-marked in the source. The labels used in GATE X-E are defined in Table \ref{tab:label_definitions}, along with examples for each.

\begin{table*}[p]
    \centering
    \begin{tabular}{p{2.85in}p{2.85in}}
    \toprule
    {\bf Description} & {\bf Example (tr $>$ en)} \\
    \midrule \midrule
    \multicolumn{2}{c}{\bf Negative/Non-AGME labels} \\
    \midrule
    \texttt{\bf source+target\_gendered\_noun} & \\
    A person is referred to by a gendered noun in both source and translations. & Git ve {\bf erkek kardeş}ine yardım et. $\rightarrow$ \newline Go and help your {\bf brother}. \\
    \midrule
    \texttt{\bf source+target\_gendered\_noun+pronoun} & \\ 
    A person referred to by a gendered noun in the source is referred to by both a gendered noun and one or more gendered pronouns in the translations. & {\bf Anne}m zaten karar{\bf ı}nı verdi. $\rightarrow$ \newline My {\bf mom} has already made {\bf her} decision. \\
    \midrule
    \texttt{\bf source\_gendered\_noun\_target\_pronoun} & \\ 
    A person is referred to by a gendered noun in the source, and one or more gendered pronouns in the translations (but not by a gendered noun). & {\bf O}, gerçek bir {\bf bilim adamı}dır. $\rightarrow$ \newline {\bf He} is a {\bf scholar} to the core. \newline (\emph{bilim adamı} indicates a male scholar) \\
    \midrule
    \texttt{\bf non-AGME-name} & \\
    A non-AGME person is referred to by name. & {\bf Umut}'un \emph{torunu} ünlü bir yazar değil mi?  $\rightarrow$ \newline {\bf Umut}'s \emph{granddaughter/grandson/grandchild} is a famous writer, isn't/aren't \emph{she/he/they}?\\
    
    \midrule \midrule
     \multicolumn{2}{c}{\bf Positive/AGME labels} \\
    \midrule
    \texttt{\bf target\_only\_gendered\_noun} & \\
    A person who is not gender-marked in the source is referred to with a gendered noun in the translations. & {\bf Yeğen}im bugün geliyor. $\rightarrow$ \newline My {\bf niece/nephew} is coming today. \\
    \midrule
    \texttt{\bf target\_only\_gendered\_pronoun} & \\
    A person who is not gender-marked in the source is referred to with a gendered pronoun in the translations. & {\bf Onun} yardımı paha biçilmezdi. $\rightarrow$ \newline {\bf Her/His/Their} help has been invaluable. \\
    \midrule
    \texttt{\bf target\_only\_gendered\_noun+pronoun} & \\
    A person who is not gender-marked in the source is referred to with both a gendered noun and gendered pronoun in the translations. & {\bf Torun}un iş{\bf ini} seviyor olmalı. $\rightarrow$ \newline Your {\bf granddaughter/grandson/grandchild} must love {\bf her/his/their} job. \\
    \midrule
    \texttt{\bf name} & \\
    An AGME is referred to by name. We treat personal names as non-gender-marking. & {\bf Beyza} akşam yemeğini bitiremedi. $\rightarrow$ \newline {\bf Beyza} wasn't able to finish {\bf her/his/their} dinner. \newline (\emph{Beyza} is typically considered a feminine name)\\
    \midrule \midrule
     \multicolumn{2}{c}{\bf Other} \\
    \midrule

    \texttt{\bf mixed} & \\
    Both positive and negative examples are present & \emph{Baba}{\bf sı} yine uçağ\emph{ını} kaçırdı. $\rightarrow$ \newline {\bf Her/His/Their} \emph{father} missed \emph {his} plane again.\\
    \midrule
    \texttt{\bf \emph{N} AGME(s)} & \\
    \emph{N} is a whole number representing the number of AGMEs in the instance. Negative examples are annotated as 0 AGMEs.& 
    0 AGME: My mother read her book. \newline
    1 AGME: {\bf She/He} ate {\bf her/his} lunch alone. \newline
    2 AGME: {\bf She/He} annoyed \emph{her/him} with 
    {\bf her/his} music.
    \\
    \bottomrule
    \end{tabular}
    \caption{{\bf Label Definitions and Examples.} Words relevant to the label are bolded or italicized in source and target. \emph{Pronoun} in these definitions includes possessive determiners \emph{her}, \emph{his}, \emph{their}.}
    \label{tab:label_definitions}
\end{table*}

\subsection{Corpus Statistics}

Counts for each label type per language can be seen in Table \ref{tab:label_counts} in appendix \ref{sec:further_details}. We present the distribution of sentence lengths in source and target languages in Figure \ref{fig:mt_bias_mitigation_2}. 

More than half of the instances for each language pair have one AGME, with around 20-30\% having no AGMEs at all. Most AGMEs are \texttt{target\_only\_gendered\_pronoun}, as there are relatively few nouns which are gendered in English but not gendered in the source languages. We see about a quarter of AGMEs are also referred to by name.

Non-AGME references will involve a gendered noun on the source, and for most of the languages about half of these also include a pronoun reference. 

%% file: latex/content/2_5_rewriting_strategy.tex
\section{Techniques for Countering Gender Bias}
\label{sec:rewriting_strat}

We propose two mitigation strategies for providing a set of gendered and gender-neutral translations. The first entails rewriting the English translation output. The second involves direct translation of the source into all possible gendered and gender-neutral variants.

\subsection{Target Rewriting}

Translation gender rewriting is the process of taking a translated source-target pair and producing alternative translations with different gender markings. It is essential, however, that gender markings in the rewrites are still compatible with gender information found in the source. 

\subsubsection{Gender Rewriting Problem Sub-Types}
\label{sec:problem_sub_types}

\begin{table*}
\centering
\begin{tabular}{llll}
\toprule
Category & Feminine & Masculine & Neutral \\
\midrule
Subject & She & He & They \\
Object & Her & Him & Them \\
Possessive Determiner & Her & His & Their \\
Possessive Pronoun & Hers & His & Theirs \\
Reflexive & Herself & Himself & Themselves \\
\bottomrule
\end{tabular}
\caption{Pronoun categories}
\label{tab:pronoun_cats}
\end{table*}

English-target rewriting problems can be broadly divided into a few subtypes, which require the use of different techniques in order to produce correct rewrites. The important features to consider are the presence or absence of intrinsically gendered nouns, such as \emph{mother}, and whether non-uniform assignment of gender in the rewrites is desired, or blanket all-female, all-male, or all-neutral assignment is acceptable.

\vspace{.25cm}
\subsubsection*{Pronoun-Only Uniform-Gender Problems}

These problems, where the only gender markers present are gendered pronouns, are the most straightforward to rewrite. By making a few assumptions and assertions, we can demonstrate that such translations can be successfully rewritten by considering only the syntax of the target sentence, and without using the source sentence. It requires a handful of deterministic pronoun string replacements, disambiguation for two ambiguous pronoun forms, and adjustment of some verb forms when their subjects change. 

Table \ref{tab:pronoun_cats} shows five categories of gendered pronouns, and in this category of rewrites, we replace pronouns found in the male and female columns with those of the same category in the desired gender. In most cases a simple lookup and replace is sufficient, but these two pronouns require disambiguation: 

\vspace{.25cm}
\hspace{1cm}\emph{her} $\rightarrow$ \emph{him/his}, \emph{them/their}

\hspace{1cm}\emph{his} $\rightarrow$ \emph{her/hers}, \emph{their/theirs}
\vspace{.25cm}

Adjustment of verb surface forms entails converting third-person singular verb forms to third-person plural when subjects are changed from \emph{she} or \emph{he} to \emph{they}, as seen in the following example:

\begin{center}
\vspace{.25cm}
\emph{He {\bf was} the oldest.}$\rightarrow$\emph{They {\bf were} the oldest.}
\vspace{.25cm}
\end{center}

The assumptions that allow a simplified approach to be successful on pronoun-only uniform-gender problems are as follows:
\begin{itemize}
\item {\bf Target Pronoun Only:} The target contains gendered pronouns, but no gendered nouns or other gender-marking modifiers (such as the adjective \emph{male}).
\item {\bf Source Gender Neutral: } Gendered pronouns do not exist in the source language.
\item {\bf Source Pronoun Only:} There are no gendered nouns or  gender-marking modifiers in the source. Note that we will later discuss a corner case where this assumption is false even when the target-pronoun-only assumption holds.
\item {\bf Limited Context:} The only context relevant to the translation is present in the source text. Therefore there is not possible that gender markings found in the target could have been implied by external context.
\item {\bf Uniform Gender:} All AGMEs mentioned in the target language should be rewritten as if they have the same gender -- all female, all male, or all gender-neutral.
\end{itemize}

Without gendered nouns (source-pronoun-only and source-gender-neutral assumptions) or external context (limited-context assumption), we are guaranteed that no gender markings found in the target were forced during translation, and thus all gender-marked individuals in the target must be AGMEs. The target-pronoun-only assumption guarantees that the \emph{only} indications of gender in the target are gendered pronouns. With the uniform-gender assumption, we must assign the same gender to all AGMEs, and so we merely need to perform surface level replacements for \emph{all} gendered pronouns with analogous pronouns of the desired output gender, as well as adjusting verb endings in the case of \emph{they} as previously described. 

Where our above assumptions hold, we can completely ignore the source sentence when translating from a gender-neutral language, and we know that any time we see gendered pronouns in the target, rewrites are appropriate, such as in the following example:

\vspace{.25cm}
  [arbitrary source] $\rightarrow$ \emph{She gave him her umbrella.}
\vspace{.25cm}

Here, under the most likely reading, the subject of \emph{gave} refers to the same person as the owner of the umbrella. This leaves us with two people, and we know that neither person's gender was indicated in the source. Therefore, it is valid to reinterpret either or both individuals' genders. 

The umbrella giver could also be male or neutral and the receiver could also be female or neutral without breaking the translation's correctness. With the uniform-gender assumption, any rewrite would have both individuals given the same gender assignment, with three possible results as follows:

\begin{table}[!h]
\tabcolsep 3pt
\centering
\begin{tabular}{ll}
\toprule
{\fontsize{10}{12}\bf Female + Female}& {\fontsize{10}{12}She gave her her umbrella.} \\ 
{\fontsize{10}{12}\bf Male + Male}&{\fontsize{10}{12}He gave him his umbrella.} \\  
{\fontsize{10}{12}\bf Neutral + Neutral}&{\fontsize{10}{12}They gave them their umbrella.}\\
\bottomrule
\end{tabular}
\end{table}

Theoretically an individual referred to with gender neutral pronouns (e.g. singular \emph{they}) could also be rewritten to use gendered pronouns, but it would then be necessary to distinguish it from plural uses of \emph{they}. This may be possible from context or by correlating it with the corresponding word used in the source sentence. However, attempting to do so requires more than the simple pronoun replacement described above.

\vspace{.25cm}
\subsubsection*{Pronoun-Only Non-Uniform Gender Problems} 

To consider rewrites where we allow different gender assignments for different AGMEs within an instance, we must relax the uniform-gender constraint. If we do so, then in addition to the three rewrites above, the following six also become possible, one of which being the original translation:

\begin{table}[!h]
\tabcolsep 3pt
\centering
\begin{tabular}{ll}
\toprule
{\fontsize{10}{12}\bf Female + Male}& {\fontsize{10}{12}She gave him her umbrella.} \\ 
{\fontsize{10}{12}\bf Female + Neutral}&{\fontsize{10}{12}She gave them her umbrella.} \\  
{\fontsize{10}{12}\bf Male + Female}&{\fontsize{10}{12}He gave her his 
umbrella.}\\
{\fontsize{10}{12}\bf Male + Neutral}& {\fontsize{10}{12}He gave them his umbrella.} \\ 
{\fontsize{10}{12}\bf Neutral + Female}&{\fontsize{10}{12}They gave her their umbrella.} \\  
{\fontsize{10}{12}\bf Neutral + Male}&{\fontsize{10}{12}They gave him their
umbrella.} \\ 
\bottomrule
\end{tabular}
\end{table}

The difficulty with relaxing this constraint is that it is now necessary to determine which pronouns refer to the same individuals as which other pronouns. This requires the ability to solve an intrasentential coreference problem, which is more difficult than the shallow surface form adjustments needed for the uniform gender case.

If we are able to identify clusters of coreferent pronouns, we can assign a desired gender to each cluster independently, and then apply the same string transformation rules per pronoun that were applied in the uniform gender case.

\vspace{.25cm}
\subsubsection*{Gendered-Noun Problems} If we relax the target-pronoun-only and source-pronoun-only assumptions and expand our scope to include translations containing gendered nouns, we encounter a host of new challenges that render the the rewriting problem significantly more difficult. The following pair of examples illustates some of those new challenges.

\vspace{.25cm}
\begin{center}
\emph{Kardeşine ziyarete gelip gelmeyeceğini sordu.}

$\big\Downarrow$ 

\emph{He asked his sister if she would visit.}
\end{center}

In this translation, both the male and female individuals are AGMEs since \emph{Kardeşine} simply denotes a \emph{sibling} without any gender specification. Therefore, there are nine possible rewrites, including the original translation: \emph{He} and \emph{his} can be optionally replaced with \emph{she/her} or \emph{they/their}, and \emph{sister/she} can be optionally replaced with \emph{brother/he} or \emph{sibling/they}.

In the next example, however, the gender of the sister is specified as female by the addition of the word \emph{Kız}, even though the default English translation is exactly the same:
\vspace{.25cm}
\begin{center}
\emph{\textbf{Kız} kardeşine ziyarete gelip gelmeyeceğini sordu.}

$\big\Downarrow$ 

\emph{He asked his sister if she would visit.}
\end{center}
\vspace{.25cm}

Here \emph{sister} must remain fixed because of the gender marking in the source. \emph{She} is also fixed because it is coreferent with the sister. Only the individual referred to by \emph{he} and \emph{his} is an AGME, so only three valid rewrites exist (including the original):

\begin{table}[!h]
\tabcolsep 3pt
\centering
\begin{tabular}{ll}
\toprule
{\fontsize{10}{12}\bf Male}& {\fontsize{10}{12}He asked his sister if she would visit.} \\ 
{\fontsize{10}{12}\bf Female}&{\fontsize{10}{12}She asked her sister if she would visit.} \\  
{\fontsize{10}{12}\bf Neutral}&{\fontsize{10}{12}They asked their sister if she would visit.}\\
\bottomrule
\end{tabular}
\end{table}

More specifically, the challenges inherent in this problem class include the following. 

\begin{itemize}

\item Some gendered nouns may need to be rewritten to opposite gender or gender-neutral forms. This requires handling of a significantly larger set of vocabulary, and there may not be a consensus on the most appropriate gendered and gender-neutral variants for some words.  
\item With the introduction of gendered nouns, gender marking becomes possible in source sentences. If a gendered noun is present in the target, it is now necessary to identify what source noun it was translated from and whether it was gendered in the source as well. We should attempt to provide alternatives for that word only if it was \emph{not} gendered in the source. This is demonstrated by the behavior of \emph{sister} in the two examples above.
\item Gendered pronouns may also be present in the translation. Whether they can be rewritten depends on whether they are coreferent with a noun that is marked for gender in the source. There may be multiple individuals referred to by pronouns, and only some of the pronouns may be coreferent with such a source-gender-marked noun, meaning that only a subset of pronouns can be rewritten. This is demonstrated by the behavior of \emph{she} in the two examples above.

\end{itemize}

Solving these problems requires both coreference resolution and alignment of nouns between the source and target, significantly increasing the complexity of a solution.

\vspace{.25cm}
\subsubsection*{Gendered Nouns Translated into Pronouns} 
In some rare cases, it is possible for a gendered noun in one of these source languages to be translated into a gendered pronoun in English. This can happen especially in contexts where distinguishing gender is essential to the expressed meaning. For example, this Turkish-English translation can be possible if the man and woman are known from prior context.:

\begin{center}
\vspace{.25cm}
Kadın televizyon izlerken adam piyano çalıyor

(\emph{The man plays piano while the woman watches tv}) 

$\big\Downarrow$ 

He plays piano while she watches tv
\vspace{.25cm}
\end{center}

Without access to gendered pronouns, using nouns like \emph{man} and \emph{woman} may be the best available means of emphasizing \emph{which} individual is doing \emph{what}, while in English simply using pronouns can be more natural. Instances of this type use the {\fontsize{10}{12}\texttt{source\_gendered\_noun\_target\_pronoun}} label. If we were to treat such instances as monolingual rewrites and perform simple pronoun replacement, we would incorrectly produce alternative translations, where gender in the translation contradicts gender information in the source text.

\subsubsection{Monolingual Rewriting for Translation}
\label{sec:monolingual_rewriting}

While GATE X-E contains instances that cover the full range of problem classes described in section \ref{sec:problem_sub_types}, we introduce a novel solution for rewriting instances in the pronoun-only uniform-gender scenario. We do include instances that contain personal names in this category, and treat them as though those names do not mark for gender.

We consider this from the viewpoint of a user who wishes to see a set of three gendered-alternative translations: all-female, all-male and all-neutral. To solve this we first produce an all-neutral rewrite, and then use a rule based solution to convert the all-neutral rewrite to gendered rewrites. If all gendered pronouns in the source have the same gender, then one of these outputs will be a duplicate of the original input translation.

As explained in section \ref{sec:problem_sub_types}, the trickiest aspects of the gender-neutral rewrite are disambiguating pronoun classes for \emph{her} and \emph{him}, and adjusting verb forms when subjects change from \emph{she/he} to \emph{they}.
Our solution leverages GPT-3.5 Turbo's ability to handle such tasks in order to generate an all-neutral rewrite. While there are a subset of instances in the pronoun-only uniform-gender class that can be solved with a fully rule-based solution (i.e. those that do not contain \emph{she, he, her, his}), we wish to explore GPT's capabilities, and so in our experiments we use it to generate neutral rewrites even for such instances.

As we have demonstrated, problems in this class can be solved with monolingual rewriting on the target alone, and so we give GPT-3.5 Turbo access to target but not source information.
We experiment with zero-shot and few-shot approaches. The zero-shot approach uses the prompt from Figure \ref{fig:zero-shot} in appendix \ref{sec:prompting_templates}. The few-shot approach expands on this prompt by adding five examples, and it can be seen in Figure \ref{fig:few-shot}.

\subsubsection*{Gender-Neutral to Gendered Rewrite}

One useful observation is that allows us to simply the remainder of the task is that, if we can correctly generate an all-neutral rewrite, a simple rule-based algorithm guarantees correct gendered feminine and masculine rewrites as well. If we refer back to \ref{tab:pronoun_cats}, we can see that all elements in the neutral column are unique, while the masculine and feminine columns each have one surface form fitting two categories. If we know that neutral pronoun refers to a single individual, we can uniquely determine which feminine or masculine pronoun is in the same category. 

We can therefore combine information from the original translation and the all-neutral rewrite to fully determine masculine and feminine rewrites. We begin with the original translation and then replace the form for each gendered pronoun according to Table \ref{tab:pronoun_cats}. The pronoun in the same position in the gender-neutral rewrite is used to determine  the pronoun category, and then we select the pronoun of the desired gender from that category, as shown in \ref{tab:original_neutral_gendered}

In practice, the neutral rewrite may have errors, so we can modify this to only look up the category row using the neutral token when the original token was \emph{her} or \emph{his}. Otherwise we can directly look up the row using the gendered token from the original translation. We use this modification in our experiments.

\subsection{Mitigation Strategies Based on Source}
Rather than rewriting English-target translations into feminine, masculine, and neutral forms, one could use the source sentence as input to create these three variants directly. This section explains how GATE X-E can be employed to assess such a system. 

The first step is to verify that the generated feminine, masculine, and neutral variants are the same, except for changes related to gender. This is a crucial step as it ensures that the meaning of the translation remains consistent, regardless of the gender. If there are differences in the translations beyond the gender-related changes, it could imply that the translation is not accurate or is introducing additional bias. After this, the generated output can be compared with the feminine, masculine, and neutral references provided in GATE X-E using contextual MT evaluation metrics.

\citet{kuczmarski2018} initially explored a source-based debiasing approach in which they enhanced a Neural Machine Translation (NMT) system to produce gender-specific translations. This was achieved by adding an additional input token at the beginning of the sentence to specify the required gender for translation (e.g., \emph{<2FEMALE> O bir doktor $\rightarrow$ She is a doctor}). However, they encountered challenges in generating masculine and feminine translations that were exactly equivalent, with the exception of gender-related changes. As a result, they later switched to a target-based rewriting approach in their subsequent work \citep{johnson2020}.

%% file: latex/content/3_experiments.tex
\section{Experiments}
\label{sec:Experiments}

\begin{table*}
\centering
\small
\begin{tabular}{llccc}
\toprule
    {\bf Language Pair} & {\bf Method} & {\bf Accuracy (\%) $\uparrow$} & {\bf BLEU $\uparrow$ } & {\bf WER $\downarrow$} \\
\midrule
\midrule
    \multirow{3}{*}{tr $\rightarrow$ en } &  \citeauthor{sun2021they} \citeyear{sun2021they}&  96.16 & {\bf 99.65 } & 0.53 \\
    & Zero-shot & 97.24 & 99.30  &  0.80\\
    & Few-shot & {\bf 98.90} &  99.55 & {\bf 0.44} \\
    \hline
    \multirow{3}{*}{hu $\rightarrow$ en} & \citeauthor{sun2021they} \citeyear{sun2021they}  & 96.14  & {\bf 99.66 } & {\bf 0.53 }\\
     & Zero-shot & 96.58 &  99.04 &  1.27\\
    & Few-shot & {\bf 97.00} &  99.03 &  1.20\\
        \hline
    \multirow{3}{*}{fi $\rightarrow$ en} & \citeauthor{sun2021they} \citeyear{sun2021they}  & 95.24 & {\bf 99.63 } &  {\bf 0.62 }\\
     & Zero-shot & 94.80 &  98.61 &  1.75\\
    & Few-shot & {\bf 96.77 } &   98.62 &  1.54 \\
        \hline
    \multirow{3}{*}{fa $\rightarrow$ en} & \citeauthor{sun2021they} \citeyear{sun2021they}  &  94.43 &  {\bf 99.57 } & {\bf 0.65 }\\
     & Zero-shot &  95.59 &  99.00  &  1.11\\
    & Few-shot & {\bf 97.84} &  99.16 &  1.01 \\
\bottomrule
\end{tabular}
\caption{{\bf Results of Gender Neutral Rewriting on the Pronoun-Only Subset of GATE X-E}. We report the performance of the rule-based system proposed by \citeauthor{sun2021they} \citeyear{sun2021they}. Additionally, we evaluate GPT-3.5 Turbo in both zero-shot and few-shot settings.}
\label{tab:gender_neutral_results}
\end{table*}
\subsection {Data Preperation}
\label{sec:data_preparation}

To produce a test set that is suitable for target-side monolingual pronoun rewriting, we first filter the data set for each language pair in GATE X-E to remove any sentences which contain a gendered noun in either source or target. This is done by removing any instances with labels containing the substring \texttt{gendered\_noun}. We also remove the handful of sentences with three or more AGMEs as they are not fully populated with all gender combination variants. 

For all remaining sentences, we test on the following rewrite scenarios: all-female$\rightarrow$all-neutral, all-female$\rightarrow$all-male, all-male$\rightarrow$all-neutral, all-male$\rightarrow$all-female. For sentences with two AGMEs, we also check rewrites from the two female/male mixed variants into \emph{all-female}, \emph{all-male} and \emph{all-neutral}. Note that we do not attempt to rewrite any neutral surface forms to  gendered ones because it is in many cases impossible to distinguish singular \emph{they} from plural without looking at the translation source.

After filtering, we are left with 627 instances for Turkish, 580 for Persian, 857 for Finnish, and 590 for Hungarian.

\subsection {  Rewriters }
\noindent{\bf Neutral rewriter Systems}:
 We consider the following rewriting systems:
\begin{enumerate}[leftmargin=*,itemsep=0.0ex, parsep=0pt] 
\item Rule-based system proposed by \citet{sun2021they}: It uses Spacy and GPT-2 to resolve ambiguity with \emph{his} and \emph{her}, and to adjust verb forms as needed. They also trained a neural model, but it was unfortunately not accessible.
\item GPT-3.5 Turbo: We evaluate GPT-3.5 Turbo on zero-shot and few-shot settings, using the prompts shown in Figures \ref{fig:zero-shot} and   \ref{fig:few-shot} in the appendix. We set the temperature $T=0$. 
\end{enumerate}
We investigated the neural model introduced by \citet{vanmassenhove-etal-2021-neutral}  as well, but were unable to reproduce results on their test data.
 
\noindent{\bf Gendered Alternatives Rewriter System:}
We employ the Gender-Neutral to Gendered Rewriting algorithm outlined in Section \ref{sec:monolingual_rewriting}.

%% file: latex/content/4_results.tex
\section{Results}
\label{sec:Results}
\begin{figure}[t]
\centering \includegraphics[width=0.9\linewidth]{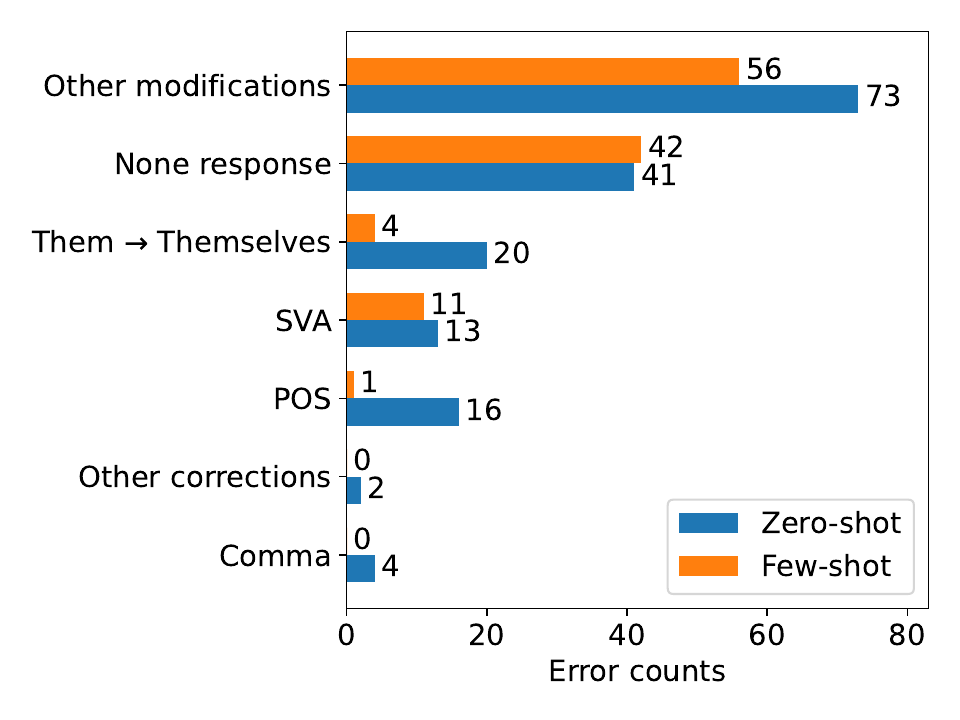}
\caption{Distribution of errors in GPT-3.5 Turbo's zero-shot and few-shot settings. The majority of errors in both settings stem from unrelated modifications and the model's 'None' response, indicating no need for gender-neutral rewriting.}
\label{fig:error_bar_plot}
\end{figure}

\subsection{Evaluation}
We report the rewriter systems' performance using BLEU \citep{papineni2002bleu}, Word Error Rate (WER), and Accuracy.

In the gender-neutral rewriting task (Table \ref{tab:gender_neutral_results}), GPT-3.5 Turbo performs better in the few-shot setting compared to the zero-shot setting. Although GPT-3.5 Turbo provides slightly higher accuracy compared to the rule-based system proposed by \citet{sun2021they}, the rule-based system performs better based on BLEU and WER. This is because GPT-3.5 Turbo makes modifications unrelated to neutral rewriting, as detailed in the error analysis section.

In the gendered-alternatives rewriting task (Table \ref{tab:gendered_results}), the zero-shot setting indicates that for resolving the \emph{her}$\rightarrow$\emph{his/him} and \emph{his}$\rightarrow$\emph{her/hers} ambiguity, gender-neutral rewrites from the zero-shot prompt are used. Similarly, the few-shot setting uses the corresponding gender-neutral outputs from the few-shot prompt. The performance of both settings is comparable.
\subsection{Error Analysis}
\begin{table*}
\centering
\small
\begin{tabular}{llccc}
\toprule
    {\bf Language Pair} & {\bf Method} & {\bf Accuracy (\%) $\uparrow$} & {\bf BLEU $\uparrow$ } & {\bf WER $\downarrow$} \\
\midrule
\midrule
    \multirow{3}{*}{tr $\rightarrow$ en } 
    & Zero-shot & {\bf 99.50 } & {\bf 99.90 } & {\bf 0.90} \\
    & Few-shot &  99.46 &  99.00 &  0.10\\
    \hline
    \multirow{3}{*}{hu $\rightarrow$ en} 
     & Zero-shot & {\bf 99.27 }&  {\bf 99.95} &  {\bf 0.08}\\
    & Few-shot &  99.20 & 99.94 &  0.09\\
        \hline
    \multirow{3}{*}{fi $\rightarrow$ en} 
     & Zero-shot & 98.41 &  {\bf 99.85 } &  0.24\\
    & Few-shot & {\bf 98.99 } &  99.80 & {\bf 0.19}\\
        \hline
    \multirow{3}{*}{fa $\rightarrow$ en} 
     & Zero-shot &  98.75 &  99.91 &  1.13\\
    & Few-shot & {\bf 99.00} & {\bf 99.93} & {\bf 0.09}\\
\bottomrule
\end{tabular}
\caption{{\bf Results of Gendered Alternatives on the Pronoun-Only Subset of GATE X-E}. Gendered alternatives are generated using the algorithm described in Section \ref{sec:monolingual_rewriting}}
\label{tab:gendered_results}
\end{table*}

Figure \ref{fig:error_bar_plot} illustrates the distribution of aggregated errors across four language pairs for GPT-3.5 Turbo in both zero-shot and few-shot settings, specifically for the task of gender-neutral rewriting. The definitions of these errors are provided in Table \ref{error_definition} in the appendix, while Table \ref{tab:error_examples} offers examples for each error label. 

In both settings, the majority of errors stem from modifications unrelated to gender-neutral rewriting and from instances where the model suggests no changes are necessary to render the input text gender-neutral. Additional examples of errors due to unrelated modifications can be found in Table \ref{tab:error_examples_other_modifications} in the appendix. The few-shot setting, however, does show an improvement in neutral rewriting errors (such as POS(part-of-speech) errors and \emph{them} being rewritten as \emph{themselves}) when compared to the zero-shot setting. 

Tables \ref{tab:zero_shot_error_counts} and \ref{tab:few_shot_error_counts} present the error distribution for each of the four languages. Upon closer examination of the Finnish data, which has the highest error rate, we found that the errors are primarily due to the longer input length. This increases the scope for modifications of the text that are unrelated to gender-neutral rewriting.

%% file: latex/content/5_related_work.tex
\section{Related Work}
\label{sec:Related_Work}

\noindent\textbf{Evaluation Benchmarks}: Although numerous studies \citep{prates2019assessing,fitria2021,ciora2021examining,ghosh2023chatgpt} have focused on translating from a weakly gendered language such as Turkish into a more strongly gendered language like English, there has been no benchmark for evaluating this phenomenon systematically. This work is the first to establish a benchmark that employs natural sentences across multiple domains and of varying lengths. The test data examines the intricacies of gendered nouns and how they influence the problem. Moreover, our research covers four languages: Turkish, Hungarian, Finnish, and Farsi.  

\noindent\textbf{Bias mitigation}: To mitigate gender bias when translating queries that are gender-neutral in the source language, Google Translate announced a feature \citep{kuczmarski2018,johnson2020} that provides gender-specific translations. This feature offers masculine and feminine translation alternatives when translating from Turkish, Finnish, Hungarian, or Persian into English. 

Both \citet{sun2021they} and \citet{vanmassenhove-etal-2021-neutral} have explored the gender-neutral rewriting of English. Their studies demonstrate that a neural model can perform this task with reasonable accuracy.

\citet{ghosh2023chatgpt} evaluates gender bias in GPT-3.5 Turbo output when translating form gender-neutral languages into English. However, to the best of our knowledge, our work is the first to leverage GPT-3.5 Turbo for mitigation, using it to achieve very high sentence-level accuracy on the pronoun-only subset of the proposed benchmark. 

%% file: latex/content/6_conclusion.tex
\section{Conclusion}

We have presented GATE X-E, a diverse dataset covering a wide range of scenarios relevant to translation gender-rewriting for English-target language pairs, covering both gendered and gender-neutral rewrites. We have discussed some intricacies of the English-target translation rewrite problems, and explained what features in a test instance lead to easier or more difficult rewrite problems. We have explored a solution using GPT-3.5 Turbo to solve a subset of these problems.

We look forward to exploring strategies that can be successful at tackling the full range of problem types covered by GATE X-E. We also hope that by making GATE X-E accessible to the broader research community, we can encourage further research on gender debiasing in the machine translation space.

%% file: latex/content/7_appendix.tex
\appendix

\section{Further details on GATE X-E}
\label{sec:further_details}

Table \ref{tab:label_counts} provides a comprehensive breakdown of corpus statistics for GATE X-E, with instance counts per language for each label.

\begin{figure*}
\centering \includegraphics[width=0.9\linewidth]{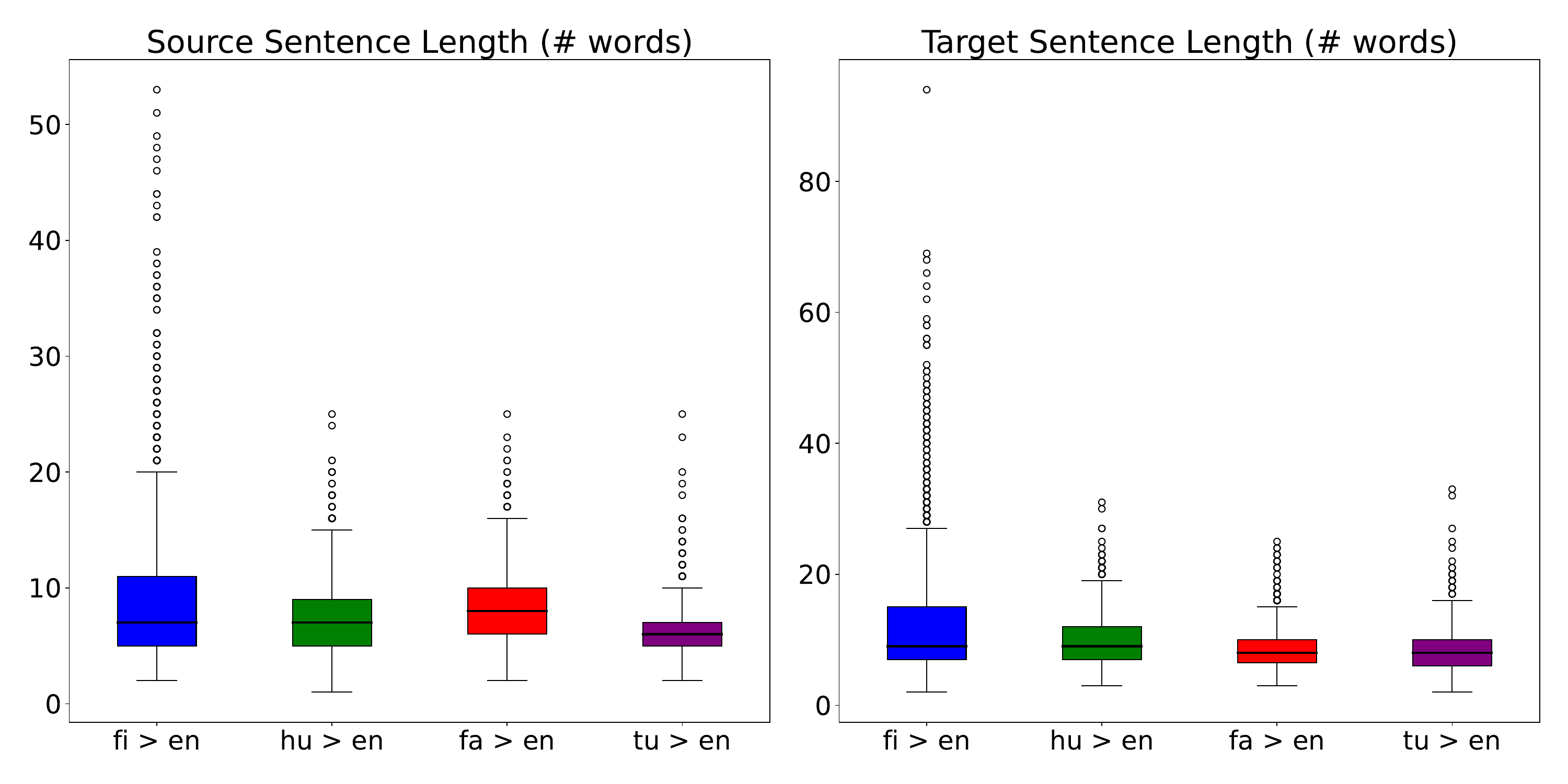}
\caption{ {\bf Boxplots representing the distribution of sentence lengths in source and target languages.} The four language pairs are Finnish to English (fi > en), Hungarian to English (hu > en), Persian to English (fa > en), and Turkish to English (tu > en). The left plot represents the source language sentence lengths, and the right plot represents the target language (English) sentence lengths. The color of each boxplot corresponds to the language pair as indicated in the legend.}
\label{fig:mt_bias_mitigation_2}
\end{figure*}

Figure \ref{fig:mt_bias_mitigation_2} presents boxplots that demonstrate the sentence length distribution on source and target for four language pairs: Finnish to English, Hungarian to English, Persian to English, and Turkish to English. The left plot shows the sentence lengths in the source languages, and the right plot displays the sentence lengths in English, the target language. The legend indicates the color corresponding to each language pair. Compared to the other three language pairs, Finnish to English contains longer sentences.

\begin{table*}
    \centering
    \begin{tabular}{lrrrr}
    \toprule
        & \bf{tr $\rightarrow$ en} & \bf{fa $\rightarrow$ en} & \bf{fi $\rightarrow$ en} & \bf{hu $\rightarrow$ en} \\
        \midrule
        \midrule
        \bf{total instance count} & 1,429 & 1,259 & 1,832 & 1,308 \\
        \midrule
        target\_only\_gendered\_noun & 142 & 118 & 159 & 95 \\ 
        target\_only\_gendered\_pronoun & 1,074 & 906 & 1,096 & 914 \\
        target\_only\_gendered\_noun+pronoun & 114 & 49 & 105 & 115 \\
        source+target\_gendered\_noun & 239 & 244 & 379 & 75 \\ 
        source+target\_gendered\_noun+pronoun & 328 & 292 & 361 & 422 \\ 
        source\_gendered\_pronoun\_target\_noun & 3 & 0 & 0 & 33 \\ 
        \midrule
        0 AGMEs & 300 & 264 & 502 & 264 \\
        1 AGME & 900 & 869 & 1,164 & 848 \\ 
        2 AGMEs & 225 & 124 & 161 & 192 \\ 
        3 AGMEs & 4 & 2 & 5 & 4 \\
        \midrule
        mixed & 271 & 263 & 237 & 262 \\
        name & 328 & 175 & 408 & 159 \\
        non-AGME-name & 32 & 5 & 136 & 16 \\
        \bottomrule
    \end{tabular}
    \caption{{\bf GATE X-E Statistics.} Sentence counts per language associated with each label.}
    \label{tab:label_counts}
\end{table*}

\section{Prompting Templates}
\label{sec:prompting_templates}

Figures \ref{fig:zero-shot} and \ref{fig:few-shot} show the GPT-3.5 Turbo zero-shot and few-shot prompts used in the gender-neutral rewriting task. We use the same prompt across all the language pairs as the task is source agnostic. 

\input{latex/figures/prompt_example_0}
\input{latex/figures/prompt_example_1}

\begin{table*}
  \centering
  \begin{tabular}{p{5cm} p{5cm} p{5cm}}
    \toprule
    Original & Gender-Neutral & Gendered Alternatives \\
    \midrule
    The teacher compared my poem with one of \textcolor{purple}{his}.
 & The teacher compared my poem with one of \textcolor{purple}{theirs}.
 & The teacher compared my poem with one of \textcolor{purple}{hers.}\\
 
    &  & The teacher compared my poem with one of \textcolor{purple}{his}.
\\
    \bottomrule
  \end{tabular}
\caption{Examples illustrating the generation of gendered alternatives using gender-neutral rewrites }
\label{tab:original_neutral_gendered}
\end{table*}

\section{Description of Error Labels}
Table \ref{error_definition} categorizes error labels in the neutral rewriting task. Errors unrelated to neutral rewriting include incorrect comma usage and other non-gender-related corrections. Neutral rewriting errors involve incorrect use of \emph{they}, subject-verb agreement errors, unnecessary changes from \emph{them} to \emph{themselves}, and irrelevant modifications. The table also notes when the rewriter suggests no changes are needed for gender neutrality.

Table \ref{tab:error_examples} provides examples of error labels in gender-neutral rewriting. Each row represents a different error label, with an input sentence, an output sentence with the error, and a reference sentence showing the correct rewrite. Errors include incorrect comma usage, wrong pronoun form, subject-verb agreement, unnecessary changes from \emph{them} to \emph{themselves}, and irrelevant modifications.

Table \ref{tab:error_examples_other_modifications} provides additional examples of errors categorized as 'Other Modifications' in the process of gender-neutral rewriting. Each row presents an output sentence with the error and a reference sentence showing the correct rewrite. The differences between the output and reference sentences are highlighted in red. 

Tables \ref{tab:zero_shot_error_counts} and \ref{tab:few_shot_error_counts} present comprehensive error analyses of GPT-3.5 Turbo's performance in gender-neutral rewriting, in both the zero-shot and few-shot settings respectively.

\begin{table*}
\centering
\begin{tabular}{|p{4cm}|l|p{8cm}|}
\hline
\textbf{Error Category} & \textbf{Error Label} & \textbf{Description} \\ \hline
\multirow{2}{=}{\parbox{4cm}{Corrections unrelated to neutral rewriting}} & Comma & Comma added or removed. \\ \cline{2-3}
 & Other corrections & Error corrections that are not related to making the text gender-neutral. \\ \hline
\multirow{4}{=}{\parbox{4cm}{Neutral Rewriting}} & POS & Wrong form of ‘they’ produced by the rewriter. \\ \cline{2-3}
 & SVA & Failure to make correct subject-verb agreement. \\ \cline{2-3}
 & Them $\rightarrow$ Themselves & Second appearance of them changed to themselves. \\ \cline{2-3}
 & 'None' response & The rewriter returns ‘None’ suggesting no changes needed to make the input gender neutral. \\ \hline
Irrelevant modifications & Other modifications & Modifications unrelated to neutral rewriting. \\ \hline
\end{tabular}
\caption{Description of the error labels}
\label{error_definition}
\end{table*}

\begin{table*}
    \centering
    \begin{tabular}{|c|p{3cm}|p{3cm}|p{4cm}|}
        \hline
        \textbf{Error Label} & \textbf{Input} & \textbf{Output} & \textbf{Reference} \\
        \hline
        Comma & Well, you surprised me!\textbf{,} Afshin said as she opened the door and saw Mary standing there. & Well, you surprised me! Afshin said as they opened the door and saw Mary standing there. & Well, you surprised me!\textbf{,} Afshin said as they opened the door and saw Mary standing there.
 \\
        \hline
        Other corrections & I \textbf{have} never heard of him before that.
 & I \textbf{had} never heard of them before that.
 & I \textbf{have} never heard of them before that. \\
        \hline
POS & The secretary noted down what \textbf{her} boss had said. & The secretary noted down what \textbf{they} boss had said.
 & The secretary noted down what \textbf{their} boss had said. \\
        \hline
SVA & \textbf{Does} she come here every week? & \textbf{Does} they come here every week?
 & \textbf{Do} they come here every week? \\
        \hline
Them $\rightarrow$ Themselves & She saw \textbf{her} play baseball. & They saw \textbf{themselves} play baseball.
 & They saw \textbf{them} play baseball. \\
        \hline
'None' response & He has no capacity to be a teacher. & none & They have no capacity to be a teacher. \\
        \hline
Other modifications& In any case, I will tell him about the critical tone \textbf{your} House has adopted on this issue. & In any case, I will tell them about the critical tone \textbf{their} House has adopted on this issue.
 & In any case, I will tell them about the critical tone \textbf{your} House has adopted on this issue. \\
        \hline
    \end{tabular}
    \caption{Examples for the error labels described in Table \ref{error_definition}}
    \label{tab:error_examples}
\end{table*}

\begin{table*}
    \centering
    \begin{tabular}{|p{7cm}|p{7cm}|}
        \hline
        \textbf{Output} & \textbf{Reference} \\
        \hline
They advised them to give up smoking, but they wouldn't listen.
 & They advised them to give up smoking, but they wouldn't listen \textcolor{red}{to them}.
 \\
        \hline
\textcolor{red}{They} w\textcolor{red}{ere} able to hold back their anger and avoid a fight.
 & \textcolor{red}{Jim} w\textcolor{red}{as} able to hold back their anger and avoid a fight.
\\
        \hline
The news that they had got\textcolor{red}{ten} injured was a shock to them.
 & The news that they had got injured was a shock to them.
 \\
        \hline
They have done it with the\textcolor{red}{ir} colleagues and the Committee of Legal Affairs.
 & They have done it with the colleagues and the Committee of Legal Affairs.
 \\
        \hline
In this respect \textcolor{red}{,} they have been very successful.
 &In this respect \textcolor{red}{I believe that} they have been very successful.
 \\
        \hline
They cannot be older than \textcolor{red}{me}.
 & They cannot be older than \textcolor{red}{I}.
 \\
        \hline
They suggested \textcolor{red}{goin}g to the theater, but there weren't any performances that night.
 & They suggested \textcolor{red}{to} g\textcolor{red}{o} to the theater, but there weren't any performances that night.
 \\
        \hline
    \end{tabular}
    \caption{More examples of errors of type 'Other Modifications'. Differences are in \textcolor{red}{red}.}
    \label{tab:error_examples_other_modifications}
\end{table*}

\begin{table*}
\centering
\small
\begin{tabular}{llccccc}
\toprule
    {\bf  Category} & {\bf Error Label} & {\bf tu } & {\bf hu  } & {\bf  fi} & {\bf  fa }\\
\midrule
\midrule
    \multirow{2}{*}{Corrections} &  Comma & 0 &  0  & 2 & 2 \\
    & Other Corrections & 0 &  0 &  0 & 2 \\
    \midrule
    \multirow{4}{*}{Neutral rewriting} & POS  & 2  &  1  &  9  & 4 \\
     &  SVA
 & 5 &  5 &  3 & 0 \\
    & Them $\rightarrow$ Themselves
 &  0 &  10 &  10 & 0 \\
     & 'None' response
 &  4 &  6 &  20 & 11 \\
        \midrule
    \multirow{1}{*}{Irrelevant Modifications} & 
      Other modifications  & 16 &  16  &   33  & 8 \\
         \midrule
    \multirow{1}{*}{Total} & 
      & 27 &  38  &   77  & 27 \\   
\bottomrule
\end{tabular}
\caption{Error analysis of GPT-3.5 Turbo's zero-shot performance in English gender-neutral rewriting task.}
\label{tab:zero_shot_error_counts}
\end{table*}

\begin{table*}
\centering
\small
\begin{tabular}{llccccc}
\toprule
    {\bf  Category} & {\bf Error Label} & {\bf tu } & {\bf hu  } & {\bf  fi} & {\bf  fa }\\
\midrule
\midrule
    \multirow{2}{*}{Corrections} &  Comma & 0 &  0  & 0 & 0 \\
    & Other Corrections & 0 &  0 &  0 & 0 \\
    \midrule
    \multirow{4}{*}{Neutral rewriting} & POS  & 0  &  0  &  0  & 1 \\
     &  SVA
 & 0 &  9 &  0 & 2 \\
    & Them $\rightarrow$ Themselves
 &  0 &  4 &  0 & 0 \\
     & 'None' response
 &  4 &  6 &  22 & 10 \\
        \midrule
    \multirow{1}{*}{Irrelevant Modifications} & 
      Other modifications  & 4 &  10  &   32  & 10 \\
         \midrule
    \multirow{1}{*}{Total} & 
      & 4 &  19  &   54  & 23 \\   
\bottomrule
\end{tabular}
\caption{Error analysis of GPT-3.5 Turbo's few-shot performance in English gender-neutral rewriting task.}
\label{tab:few_shot_error_counts}
\end{table*}

%% file: latex/figures/prompt_example_0.tex
\definecolor{lightgray}{RGB}{240,240,240} %
\renewcommand{\sfdefault}{cmss} 
\begin{figure*}
\begin{tcolorbox}[
    enhanced,
    colback=lightgray,
    colframe=black, 
    boxrule=0.2mm, 
    top=1mm, 
    bottom=1mm, 
    left=1mm, 
    right=1mm,
    fontupper=\sffamily\small,
    ]    
\begin{small}
\noindent
Change all gendered pronouns to use singular "they" instead. Don't modify anything else : \{input\_text\}
\noindent
\end{small}
\end{tcolorbox}
\caption{Zero-shot prompt template utilized in GPT-3.5 Turbo experiments.}
\label{fig:zero-shot}
\end{figure*}

%% file: latex/figures/prompt_example_1.tex
\definecolor{lightgray}{RGB}{240,240,240} %
\renewcommand{\sfdefault}{cmss} 
\begin{figure*}
    
\begin{tcolorbox}[
    enhanced,
    colback=lightgray,
    colframe=black, 
    boxrule=0.2mm, 
    top=1mm, 
    bottom=1mm, 
    left=1mm, 
    right=1mm,
    fontupper=\sffamily\small,
    ]    
\begin{small}
\noindent
Change all gendered pronouns to use singular "they" instead. Don't modify anything else.\\

input : His bike is better than mine.\\
gender neutral variant : Their bike is better than mine.\\\\
input : Jack bores me with stories about her trip.\\
gender neutral variant: Jack bores me with stories about their trip.\\\\
input : He kissed him goodbye and left, never to be seen again.\\
gender neutral variant : They kissed them goodbye and left, never to be seen again.\\\\
input : Is she your teacher?\\
gender neutral variant : Are they your teacher?\\\\
input : Anime director Satoshi Kon died of pancreatic cancer on August 24, 2010, shortly before her 47th birthday.\\
gender neutral variant : Anime director Satoshi Kon died of pancreatic cancer on August 24, 2010, shortly before their 47th birthday.\\\\
input : \{input\_text\}\\
gender neutral variant :

\noindent
\end{small}
\end{tcolorbox}
\caption{Few-shot prompt template utilized in GPT-3.5 Turbo experiments.}
\label{fig:few-shot}
\end{figure*}